%% file: eacl2023.tex
\pdfoutput=1

\documentclass[11pt]{article}

\usepackage{eacl2023}

\usepackage{times}
\usepackage{latexsym}
\usepackage{booktabs}
\usepackage{multirow}
\usepackage{multicol}
\usepackage{amsmath}
\usepackage{tabularx}
\usepackage{algorithm}
\usepackage{algpseudocode} 
\usepackage{xspace}
\usepackage{booktabs}
\usepackage{graphicx}
\usepackage{ amssymb }
\usepackage{xcolor}
\graphicspath{ {./Source/graphics/} }

\usepackage[T1]{fontenc}

\usepackage[utf8]{inputenc}

\usepackage{microtype}

\newcolumntype{x}[1]{>{\centering\arraybackslash\hspace{0pt}}p{#1}}

\usepackage{inconsolata}

\title{Unsupervised Improvement of Factual Knowledge in Language Models}

\author{Nafis Sadeq\thanks{\ \ This work was performed during the first author’s internship at Intuit.}$^{\ \ 1}$, Byungkyu Kang$^{ 2}$, Prarit Lamba$^{ 2}$, Julian McAuley$^{ 1}$ \\
   Intuit$^{2}$ \and UC San Diego$^{1}$  \\
    \texttt{\{nsadeq,jmcauley\}@ucsd.edu} \\
    \texttt{\{Jay\_Kang,Prarit\_Lamba\}@intuit.com}}

\begin{document}
\maketitle

\input{Source/abstract}
\input{Source/introduction}

\input{Source/related_work}

\input{Source/methodology}
\input{Source/experiments}
\input{Source/conclusion}
\input{Source/acknowledgement}
\input{Source/limitations}
\input{Source/ethical}

\bibliography{anthology,custom}
\bibliographystyle{acl_natbib}

\input{Source/appendix}

\end{document}

%% file: Source/abstract.tex
\begin{abstract}
Masked language modeling (MLM) plays a key role in pretraining large language models. But the MLM objective is often dominated by high-frequency words that are sub-optimal for learning factual knowledge. In this work, we propose an approach for influencing MLM pretraining in a way that can improve language model performance on a variety of knowledge-intensive tasks. We force the language model to prioritize informative words in a fully unsupervised way. Experiments demonstrate that the proposed approach can significantly improve the performance of pretrained language models on tasks such as factual recall, question answering, sentiment analysis, and natural language inference in a closed-book setting.
\end{abstract}

%% file: Source/introduction.tex
\section{Introduction}
Pretrained language models (PLMs) such as BERT~\cite{devlin2018bert}, RoBERTa~\cite{liu2019roberta}, BART~\cite{lewis2019bart}, T5~\cite{2020t5} use a Masked Language Modeling (MLM) objective during pretraining. %
However, a traditional MLM objective may not be optimal for knowledge-intensive tasks~\cite{peters2019knowledge}. It has been shown that language models can benefit from incorporating knowledge within the training objective in the form of entity embeddings~\cite{peters2019knowledge,zhang2019ernie}, knowledge retriever~\cite{guu2020retrieval}, knowledge embedding~\cite{wang2021kepler,colake} or augmented pretraining corpora created from Knowledge Graphs~\cite{agarwal-etal-2021-knowledge}. Despite their effectiveness, these approaches rely on existing knowledge bases and entity embeddings to incorporate knowledge within the training objective. These resources are expensive to construct and may not be available for all languages and domains~\cite{huang-etal-2022-multilingual}. 

In this work, we propose a pretraining approach that can achieve better performance on knowledge-intensive tasks without using any existing knowledge base. We combine two key strategies to influence MLM objective. Firstly, the tokens with higher informative relevance should be masked more frequently~\cite{sadeq2022informask}. Secondly, mistakes on informative tokens should be penalized more severely. %
The informative relevance of the tokens can be computed efficiently with a one-pass computation on the pretraining corpora. Experiments demonstrate that the proposed training strategy can help the language model achieve better performance on the factual knowledge recall benchmark LAMA~\cite{petroni-etal-2019-language}, extractive question answering (QA) benchmark SQuAD~\cite{rajpurkar-etal-2016-squad,rajpurkar-etal-2018-know}, prompt based sentiment analysis and natural language inference (NLI) tasks in AutoPrompt~\cite{shin-etal-2020-autoprompt}.

The key contribution of this work is proposing a completely unsupervised stand-alone MLM pretraining objective for language models that can significantly improve performance on knowledge-intensive tasks. Unlike prior works in the area, our method does not require existing knowledge bases to incorporate knowledge during pretraining. We make the code publicly available. \footnote{The code is available at \url{https://github.com/intuit/wMLM.git} }

%% file: Source/related_work.tex
\section{Related Work}

\paragraph{PLMs as knowledge bases} It has been shown that large-scale PLMs such as BERT can be used as a knowledge base~\cite{petroni-etal-2019-language,petroni2020context}. Prior works have focused on factual knowledge with regards to generative PLMs~\cite{liu2021gpt}, multilingual setting~\cite{jiang-etal-2020-x}, entities and query types~\cite{heinzerling-inui-2021-language}, fact checking~\cite{lee-etal-2020-language}.

\paragraph{Designing better prompts}  \citet{jiang-etal-2020-know} propose mining-based and paraphrasing-based methods for automatically generating prompts for improved factual recall performance. A similar approach is explored by \citet{zhong-etal-2021-factual,haviv-etal-2021-bertese,qin2021learning}. \citet{shin-etal-2020-autoprompt} propose an approach for automatically creating MLM prompts for a diverse range of tasks such as sentiment analysis, natural language inference, relation extraction, etc.

\paragraph{Knowledge integration during pretraining} \citet{peters2019knowledge} use entity embeddings from existing knowledge bases and incorporate an entity linking loss jointly with an MLM loss to improve the factual recall performance of BERT. Similarly, \citet{zhang2019ernie,wang2021kepler,fevry-etal-2020-entities,colake,liu2020kbert} use entity representations or knowledge representation from existing knowledge bases to incorporate knowledge into the PLM. %
\citet{guu2020retrieval} jointly pretrain a knowledge retriever along with a language modeling objective for knowledge integration. \citet{agarwal-etal-2021-knowledge} synthesize a text corpus from existing knowledge bases and use that during pretraining. \citet{sun2019ernie} use entity-level and phrase-level knowledge masking during training. %

\paragraph{Knowledge modification after pretraining} \citet{de-cao-etal-2021-editing,zhu2020modifying} use constraint optimization for editing existing world knowledge within PLMs with minimal impact on the rest of the factual knowledge. Similarly, \citet{verga-etal-2021-adaptable} develop a fact injection language model architecture that allows easy integration of existing knowledge bases into PLMs without additional pretraining.

%% file: Source/methodology.tex
\section{Methodology}
We use MLM objective for pretraining, similar to prior works~\cite{devlin2018bert,liu2019roberta,lewis2019bart,2020t5}. Given a sequence of tokens $Z$, a subset of tokens $X \subset Z $ is randomly sampled for replacement ($|X| / |Z| \approx 0.15$ in \citet{devlin2018bert}). For the replacement candidates in $X$, 80\% of the time the replacement is done with a special token \texttt{[MASK]}, 10\% of tokens are replaced with a random token, and the other 10\% of candidates are left unchanged \cite{devlin2018bert,liu2019roberta,joshi2020spanbert}. The task of the model during pretraining is to predict the original tokens from the modified input sequence. For a set of replaced tokens $X (x_1,x_2,...,x_N)$ and their corresponding output tokens $Y (y_1,y_2,...,y_N)$,
the loss $\mathcal{L}_{MLM}$ is computed as follows:

\vspace{-1em}
\begin{equation}
   \mathcal{L}_{MLM} = - \sum\limits_{i=1}^N \log\frac{e^{x_{i,y_i}}}{\sum\limits_{v \in V} e^{x_{i,v}}}
\end{equation}

\begin{figure}
  \centering
    \includegraphics[width=\columnwidth]{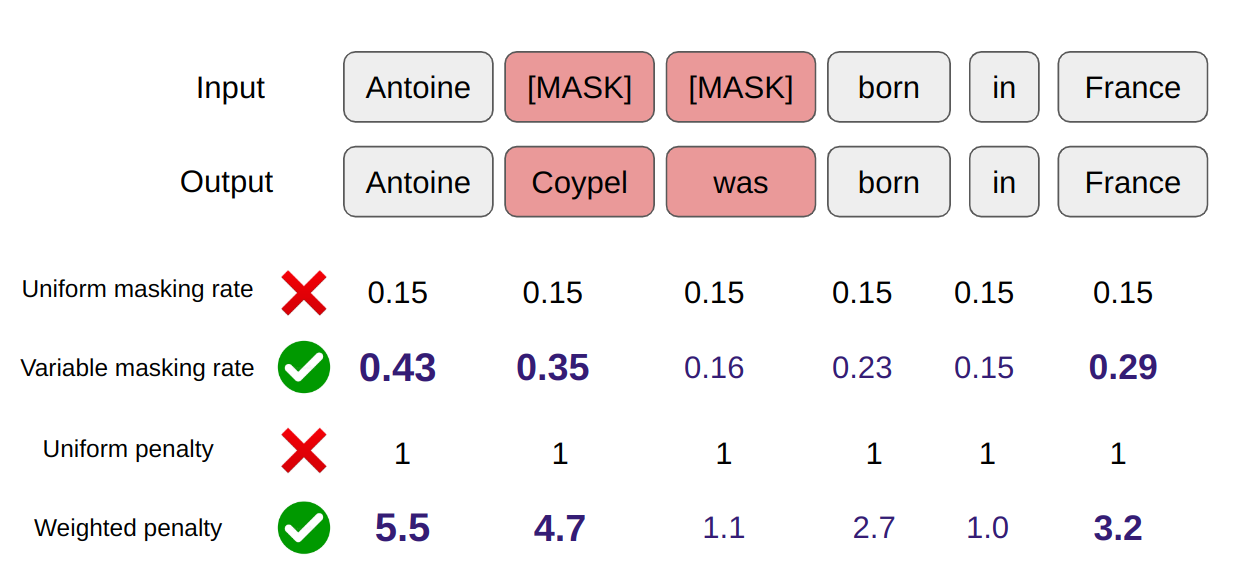}
  \caption{Simplified illustration of variable masking rate and weighted penalty}
  \label{fig:weight}
  \vspace{-1em}
\end{figure}

Here, $x_{i,j}$ is the logit produced for output candidate $j$ given input $x_i$ and $V$ is the vocabulary set. In traditional MLM loss computation, a uniform penalty is applied for all tokens within the vocabulary. In our work, we try to influence the MLM objective during pretraining to incorporate more factual knowledge. We differ from traditional MLM pretraining in two ways: \textbf{(a)} Instead of masking all tokens with equal probability, we allow some tokens to be masked more frequently if they have higher informative relevance, \textbf{(b)} We use weighted cross entropy loss to penalize mistakes on some tokens more severely if they have higher informative relevance. Simple illustrations of these two concepts are shown in Figure~\ref{fig:weight}. We compute the loss as follows:

\vspace{-1em}
\begin{equation}
\label{eq:wmlm}
    \mathcal{L}_{MLM} = - \sum\limits_{i=1}^N  w_{y_i}\log\frac{e^{x_{i,y_i}}}{\sum\limits_{v \in V} e^{x_{i,v}}}
\end{equation}

$w_{y_i}$ is a penalty weight specific to a particular output token $y_i$. The magnitude of the weight is chosen based on the informative relevance of the tokens. A demonstration of this weighting is shown in Figure~\ref{fig:weight}. %
Each token in the language model vocabulary has a unique masking rate and penalty weight associated with it. These values can be computed with a one-pass computation before pretraining.

\begin{figure}
  \centering
    \includegraphics[width=\columnwidth]{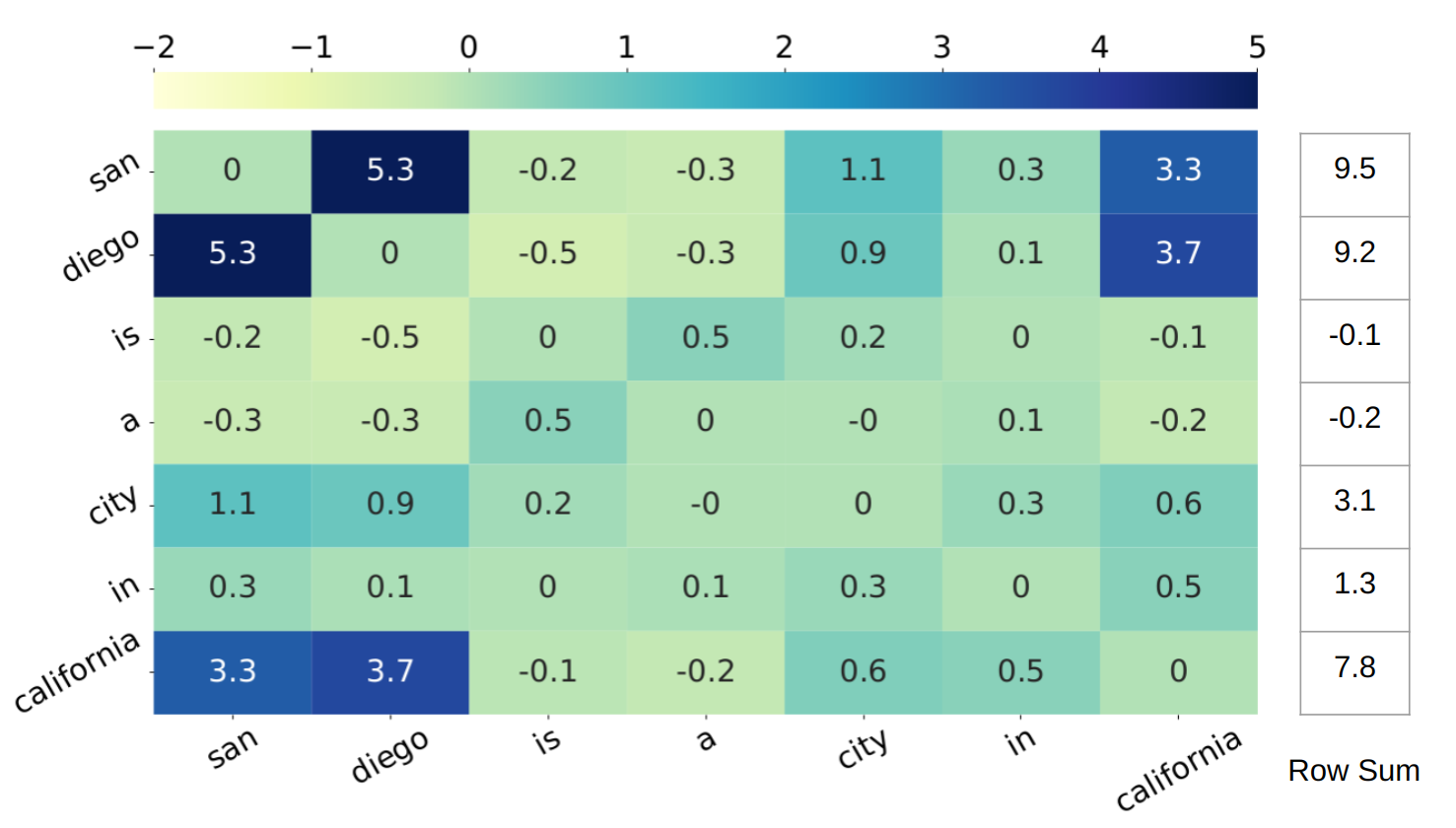}
  \caption{Informative relevance of tokens in a particular document, by computing row-wise summation of the PMI matrix of all token pairs}
  \label{fig:matrix_sum}
  \vspace{-1em}
\end{figure}

In this context, the informative relevance of tokens represents how important a particular token is with regard to the factual knowledge%
. Tokens that are more important for factual knowledge (e.g.~named entities) are expected to have a higher informative relevance. We use Pointwise Mutual Information (PMI~\cite{fano1961transmission}) to compute informative relevance in an unsupervised manner. We hypothesize that words that have high PMI with their neighboring words tend to have higher informative relevance. Firstly, we compute word co-occurrence statistics for the pretraining corpus within a skip-gram window. Secondly, PMI between all word pairs within the vocabulary is computed. Thirdly, we consider the pairwise PMI between all words within a particular document in the form of a matrix (as shown in Figure~\ref{fig:matrix_sum}), so that the row-wise sum in that matrix reflects the token-specific informative relevance within that document. Then informative relevance for a token is averaged across the corpus. Finally, the computed values are normalized and converted to token-specific masking rates and token-specific penalty weights. Those masking rates are used to create masked inputs and the penalty weights are then incorporated during MLM loss computation, as shown in Equation \ref{eq:wmlm}.

%% file: Source/experiments.tex
\section{Experiments}
\label{experiments}
\subsection{Pretraining Setup}
\label{subsec:trainsetup}
We use the Wikipedia corpus available in Hugging Face~\cite{Hugg} for pretraining, using a wordpiece tokenizer with a vocabulary size of 100k. The vocabulary size is chosen to ensure the inclusion of most entities. Word co-occurrence statistics are computed using a skip-gram window size of 10. The size of the matrix that holds the PMI between words is $100k\times100k$. The one-pass computation involving informative relevance of tokens takes around two hours and requires 11 GB of memory. The masking rate for individual tokens varies between 15\%-50\%, depending on their informative relevance. The average masking rate for all tokens is 19\%. The penalty weights for tokens are normalized within the range $[1,5]$. Training is done with Hugging Face Transformers~\cite{hftransformer} on an AWS p3.8xlarge machine with 4 Nvidia V100 GPUs. Our model architecture is similar to BERT-base~\cite{devlin2018bert} with 12 layers and a hidden dimension of 768. The overall batch size is 128 with a learning rate of 5e-5 and an AdamW optimizer~\citep{adamw}. Training is done for 10 epochs with a maximum document length of 128. Unlike BERT~\cite{devlin2018bert}, we do not use the next sentence prediction objective during pretraining. Additionally, the increased masking rate and penalty weight only apply to whole-word tokens. For the subword tokens, we use the minimum masking rate of 15\% and penalty weight of 1.

\begin{table*}[]
\centering
\resizebox{0.95\textwidth}{!}{
\begin{tabular}{lx{2cm}x{1.8cm}x{1.8cm}x{1.8cm}|x{2cm}x{2cm}x{2cm}}
\toprule
\multirow{2}{*}{Model}  & \multicolumn{4}{c}{LAMA~\cite{petroni-etal-2019-language}} & \multicolumn{3}{|c}{AutoPrompt~\cite{shin-etal-2020-autoprompt}} \\
\cmidrule{2-8}
 & ConceptNet & GoogleRE & SQuAD & TREx & SST2 & NLI (3 way) & NLI (2 way) \\
\midrule
BERT\textsubscript{uu}& 0.114 & 0.281 & 0.156 & 0.578 & 0.651 & 0.397 & 0.620 \\
BERT\textsubscript{uw}& 0.120 & 0.289 & 0.169 & 0.592 & 0.655 & 0.439 & 0.676 \\
BERT\textsubscript{vu}& 0.129 & 0.292 & 0.175 & 0.616 & 0.700 & 0.457 & 0.697 \\
BERT\textsubscript{vw}& \textbf{0.134} & \textbf{0.298} & \textbf{0.187} & \textbf{0.625} & \textbf{0.704} & \textbf{0.481} & \textbf{0.711}\\
\bottomrule
\end{tabular}
}
\caption{Factual Recall performance on LAMA, Sentiment Analysis and Natural Language Inference on AutoPrompt. The metrics used for LAMA and AutoPrompt are Mean Reciprocal Rank (MRR) and Accuracy respectively.}
\label{tab:lamaauto}
\end{table*}

\begin{table*}[]
\centering
\resizebox{\textwidth}{!}{
\begin{tabular}{lx{1.6cm}x{1.6cm}|x{1cm}x{1cm}x{1cm}x{1cm}x{1cm}x{1cm}x{1cm}x{1.1cm}x{1.1cm}}
\toprule
\multirow{2}{*}{Model} & \multicolumn{2}{c|}{SQuAD} & \multicolumn{9}{c}{GLUE~\cite{wang2018glue}} \\
\cmidrule{2-12}
 & v1~\shortcite{rajpurkar-etal-2016-squad} & v2~\shortcite{rajpurkar-etal-2018-know} & CoLA & SST2 & MNLI & QNLI & QQP & STSB & RTE & WNLI & MRPC \\
 \midrule
BERT\textsubscript{uu}& 69.96 & 83.22 & \textbf{31.06} & 88.30 & 79.42 & 87.72 & 89.77 & 85.41 & \textbf{66.43} & 42.25 & 87.78 \\
BERT\textsubscript{uw} & 71.17 & 84.17 & 28.55 & 89.11 & 79.82 & 87.15 & 89.59 & 85.70 & 58.84 & 49.30 & 87.93 \\
BERT\textsubscript{vu}& 71.17 & 85.07 & 29.11 & 89.79 & 80.02 & 88.21 & \textbf{90.10} & 85.60 & 61.37 & 54.93 & 88.29 \\
BERT\textsubscript{vw}& \textbf{72.61} & \textbf{85.28} & 28.93 & \textbf{89.91} & \textbf{80.25} & \textbf{88.49} & 89.82 & \textbf{85.82} & 59.93 & \textbf{56.34} & \textbf{88.32}\\
\bottomrule
\end{tabular}
}
\caption{Performance on SQuAD and GLUE development set. For SQuAD, we report the F1 score. We report the Matthews correlation for CoLA, Pearson correlation for STSB, and accuracy for other GLUE tasks. The fine-tuning parameters for SQuAD and GLUE can be found in Appendix~\ref{app:finetuning}.}
\label{tab:squadglue}
\end{table*}

\subsection{Evaluation Benchmarks}
We use LAMA knowledge probes~\cite{petroni-etal-2019-language} for evaluating the factual recall performance of the model. LAMA has around 70k samples across 46 factual relations. To evaluate the performance on extractive QA, we use SQuAD v1 and v2~\citep{rajpurkar-etal-2016-squad,rajpurkar-etal-2018-know}. For zero-shot performance evaluation on closed-book QA, we use the SQuAD portion from LAMA~\cite{petroni-etal-2019-language}. For closed-book sentiment analysis and NLI, we use SST2 and NLI probes from  AutoPrompt~\cite{shin-etal-2020-autoprompt}. We also report the performance of the models on GLUE~\cite{wang2018glue}.

\subsection{Baselines}
We train four models using the same corpus, tokenizer and hyper-parameter setting mentioned in Section~\ref{subsec:trainsetup}: \textbf{(a)} BERT\textsubscript{uu}: Similar to \citet{devlin2018bert}, it uses a uniform masking rate and uniform penalty across tokens. This is our baseline. \textbf{(b)} BERT\textsubscript{uw}: uses a uniform masking rate and weighted penalty. \textbf{(c)} BERT\textsubscript{vu}~\cite{sadeq2022informask}: uses a variable masking rate across tokens and uniform penalty. \textbf{(d)} BERT\textsubscript{vw}: This is our proposed approach that combines both a variable masking rate and weighted penalty across different tokens.

\subsection{Results and Discussion}

\paragraph{Factual Recall and Zero-shot QA}
The model using the proposed pretraining approach (BERT\textsubscript{vw}) significantly outperforms the baseline (BERT\textsubscript{uu}) on factual recall tasks in LAMA (shown in Table~\ref{tab:lamaauto}). The relative improvement of Mean Reciprocal Rank (MRR) over the baseline is 17.5\%, 6\%, and 8.1\% for ConceptNet, GoogleRE, and TREx respectively. The SQuAD portion of the LAMA benchmark is a set of zero-shot QA samples adapted in a closed-book template. In this task, we achieve 19.9\% relative improvement over the baseline. 

Case studies on factual recall are shown in Table~\ref{tab:casestudy}. There are two key observations in these case studies. Firstly, the proposed model (BERT\textsubscript{vw}) is more likely to rank the ground truth label higher during knowledge probes. This helps the model achieve better overall MRR. Secondly, the proposed model is more likely to produce specific words given a particular context when the baseline is only producing generic words. For example, when we use the prompt `During Super Bowl 50 the \texttt{{[}MASK{]}} gaming company debuted their ad for the first time', the top three candidates from the baseline model are comparatively common words such as `computer', `electronic', and `American'. But the proposed model is able to produce more specific words associated with three gaming companies (`Nintendo, `Walt', and `Atari'), including the correct answer `Nintendo'. Similar observation can be made with the probe `The organization that runs the satellite that measured dust that landed on the Amazon is \texttt{{[}MASK{]}}', where the proposed model makes specific predictions with the given context, such as `NASA', `Brazil' and `Amazon'. But the baseline can only produce generic words like `unknown', `the', and `unclear'.

\begin{table*}[!ht]
\resizebox{\textwidth}{!}{
\begin{tabular}{x{8cm}x{2cm}x{2.5cm}x{1cm}x{2.5cm}x{1cm}}
\toprule
\multirow{2}{*}{Input} & \multirow{2}{*}{Ground Truth} & \multicolumn{2}{c}{BERT\textsubscript{uu}~\cite{devlin2018bert}} & \multicolumn{2}{c}{BERT\textsubscript{vw} (proposed)} \\ 
\cmidrule{3-6} 
 &  & Prediction & Score & Prediction & Score \\ 
\midrule
\multirow{3}{8cm}{To emphasize the 50th anniversary of the Super Bowl the \texttt{{[}MASK{]}} color was used.} & \multirow{3}{*}{gold} & yellow & 0.17 & \textbf{gold} & 0.09 \\
 &  & red & 0.13 & rainbow & 0.06 \\
 &  & green & 0.12 & orange & 0.06 \\
 \midrule
\multirow{3}{8cm}{During Super Bowl 50 the \texttt{{[}MASK{]}} gaming company debuted their ad for the first time.} & \multirow{3}{*}{nintendo} & computer & 0.06 & \textbf{nintendo} & 0.05 \\
 &  & electronic & 0.05 & walt & 0.04 \\
 &  & american & 0.03 & atari & 0.04 \\
 \midrule
\multirow{3}{8cm}{A teacher is most likely teaching at a \texttt{{[}MASK{]}}.} & \multirow{3}{*}{school} & university & 0.61 & \textbf{school} & 0.40 \\
 &  & \textbf{school} & 0.26 & university & 0.34 \\
 &  & college & 0.03 & seminary & 0.09 \\
 \midrule
\multirow{3}{8cm}{Photosynthesis releases \texttt{{[}MASK{]}} into the Earth's atmosphere.} & \multirow{3}{*}{oxygen} & sunlight & 0.13 & \textbf{oxygen} & 0.21 \\
 &  & photosynthesis & 0.09 & carbon & 0.12 \\
 &  & light & 0.09 & sunlight & 0.06 \\
 \midrule
\multirow{3}{8cm}{The organization that runs the satellite that measured dust that landed on the Amazon is \texttt{{[}MASK{]}} .} & \multirow{3}{*}{nasa} & unknown & 0.11 & \textbf{nasa} & 0.06 \\
 &  & the & 0.03 & brazil & 0.05 \\
 &  & unclear & 0.03 & amazon & 0.02 \\
 \midrule
\multirow{3}{8cm}{Income inequality began to increase in the US in the \texttt{{[}MASK{]}}.} & \multirow{3}{*}{1970s} & 1960s & 0.21 & \textbf{1970s} & 0.14 \\
 &  & 1980s & 0.18 & 1960s & 0.13 \\
 &  & \textbf{1970s} & 0.17 & 1980s & 0.12 \\
 \midrule
\multirow{3}{8cm}{He moved to \texttt{{[}MASK{]}} at age 16 to complete his high school studies and obtained his Japanese citizenship in 1995.} & \multirow{3}{*}{japan} & tokyo & 0.42 & \textbf{japan} & 0.19 \\
 &  & \textbf{japan} & 0.21 & tokyo & 0.18 \\
 &  & yokohama & 0.03 & hawaii & 0.06 \\
 \midrule
\multirow{3}{8cm}{The Crimes Act 1914 is a piece of Federal legislation in \texttt{{[}MASK{]}}.} & \multirow{3}{*}{australia} & canada & 0.39 & \textbf{australia} & 0.12 \\
 &  & \textbf{australia} & 0.07 & tennessee & 0.09 \\
 &  & england & 0.03 & canada & 0.09 \\
 \midrule
\multirow{3}{8cm}{She is also member of the Helsinki City Council and the chairperson of the local party organisation in \texttt{{[}MASK{]}}.} & \multirow{3}{*}{helsinki} & finland & 0.52 & \textbf{helsinki} & 0.76 \\
 &  & \textbf{helsinki} & 0.38 & finland & 0.18 \\
 &  & espoo & 0.01 & espoo & 0.03 \\
 \midrule
\multirow{3}{8cm}{Mark Schwahn (born July 5, 1966) is an American \texttt{{[}MASK{]}}, director and producer.} & \multirow{3}{*}{screenwriter} & actor & 0.66 & \textbf{screenwriter} & 0.53 \\
 &  & \textbf{screenwriter} & 0.14 & writer & 0.21 \\
 &  & writer & 0.13 & actor & 0.16 \\
 \bottomrule
\end{tabular}
}
\caption{Case Study from factual recall samples from LAMA~\cite{petroni-etal-2019-language}}
\label{tab:casestudy}
\end{table*}

\paragraph{Closed-book Sentiment Analysis and NLI}
We use AutoPrompt~\cite{shin-etal-2020-autoprompt} to evaluate the closed-book sentiment analysis and NLI performance of the system. AutoPrompt provides a way to convert certain NLP tasks into a template-based probing format. The advantage of this type of prompting is that it allows us to exploit the factual knowledge within language models without the limitations of fine-tuning~\cite{singh-etal-2020-bertnesia}. The prompt contains the input, a placeholder for the answers, and a span of trigger words (prompt templates shown in Appendix~\ref{app:autoprompt}). The trigger words are tuned using the training dataset and then subsequently used during evaluation. The proposed system achieves 8.1\%, 21.1\%, and 14.7\% relative improvement in accuracy over the baseline in sentiment analysis, 3-way NLI, and 2-way NLI respectively (Table~\ref{tab:lamaauto}).

\paragraph{Fine-tuning vs Prompt-tuning}
Our proposed model achieves better performance compared to the baseline when fine-tuned on the extractive QA benchmark SQuAD~\cite{rajpurkar-etal-2016-squad,rajpurkar-etal-2018-know} and text classification benchmark GLUE~\cite{wang2018glue}. It outperforms the baseline on both SQuAD v1 and v2 tasks and seven out of nine GLUE tasks (shown in Table~\ref{tab:squadglue}). However, %
the relative performance improvement with fine-tuning is not as significant as factual recall, zero-shot QA, or prompt-tuning scenarios. The reason behind this may be explained by the findings of \citet{singh-etal-2020-bertnesia}. The main strength of our approach is the ability to store more factual knowledge during pretraining. However, \citet{singh-etal-2020-bertnesia} have shown that the factual knowledge learned during pretraining may be lost during fine-tuning, limiting the advantage of our proposed system. On the other hand, relational probing, zero-shot QA, and prompt-tuning-based NLP tasks can exploit the additional knowledge of our model more effectively, leading to much better performance.

\paragraph{Ablation Study}
We investigate how much performance improvement is due to the variable masking rate as opposed to the weighted penalty during MLM pretraining. This can be found by comparing BERT\textsubscript{uw} with BERT\textsubscript{vu} (Table~\ref{tab:lamaauto} and \ref{tab:squadglue}). In most cases, we find that a variable masking rate performs slightly better than a weighted penalty.

%% file: Source/conclusion.tex
\section{Conclusion}
In this work, we propose a pretraining strategy that can be effective in storing factual knowledge within language models. The additional knowledge helps the model outperform previous approaches on a variety of knowledge-intensive NLP tasks, such as factual recall, zero-shot QA, closed-book sentiment analysis, and natural language inference. Our model also achieves better performance when fine-tuned on SQuAD and GLUE tasks. In the future, we aim to extend our work for text-to-text pretrained models such as T5~\cite{2020t5}.

%% file: Source/acknowledgement.tex
\section*{Acknowledgements}
This work was partially supported by the Intuit University Collaboration Program grant. We thank anonymous reviewers for providing their valuable feedback on this work.

%% file: Source/limitations.tex
\section*{Limitations}
One limitation of the proposed system is that it under-performs compared to the baseline in some fine-tuning tasks, such as CoLA (Table~\ref{tab:squadglue}). The proposed training objective reduces the importance of stopwords in the pretraining objective. This may have a negative impact on performance in tasks where the syntax is important. More investigation is needed to understand and mitigate this issue.

%% file: Source/ethical.tex
\section*{Ethics Statement}
 A potential concern for the proposed system is that this training strategy may amplify the existing toxic behavior or bias of the language model if the related keywords get prioritized in the training objective. Reducing the toxic or biased behaviors of the proposed model can be an interesting research direction for future work.

%% file: Source/appendix.tex
\appendix
\onecolumn

\newpage
\section{Performance on LAMA by Relation}
\label{app:lama}

\begin{table*}[!ht]
\centering
\resizebox{0.65\textwidth}{!}{
\begin{tabular}{llrrrr}
\toprule
Domain & Dataset & BERT\textsubscript{uu} & BERT\textsubscript{uw} & BERT\textsubscript{vu} & BERT\textsubscript{vw} \\
\midrule
ConceptNet & test & 0.114 & 0.120 & 0.129 & \textbf{0.134} \\
GoogleRE & dateOfBirth & 0.099 & 0.109 & 0.111 & \textbf{0.113} \\
GoogleRE & placeOfBirth & 0.456 & 0.459 & 0.461 & \textbf{0.465} \\
GoogleRE & placeOfDeath & 0.288 & 0.300 & 0.305 & \textbf{0.315} \\
Squad & test & 0.156 & 0.169 & 0.175 & \textbf{0.187} \\
TREx & P1001 & 0.779 & 0.770 & 0.793 & \textbf{0.798} \\
TREx & P101 & 0.442 & 0.468 & 0.501 & \textbf{0.514} \\
TREx & P103 & 0.822 & 0.834 & \textbf{0.838} & 0.836 \\
TREx & P106 & 0.642 & 0.653 & \textbf{0.675} & 0.664 \\
TREx & P108 & 0.491 & 0.526 & 0.538 & \textbf{0.556} \\
TREx & P127 & 0.586 & 0.615 & 0.620 & \textbf{0.636} \\
TREx & P1303 & 0.380 & 0.427 & 0.433 & \textbf{0.472} \\
TREx & P131 & 0.690 & 0.702 & 0.741 & \textbf{0.750} \\
TREx & P136 & 0.595 & 0.629 & 0.651 & \textbf{0.675} \\
TREx & P1376 & 0.747 & 0.761 & 0.783 & \textbf{0.792} \\
TREx & P138 & 0.633 & 0.640 & 0.656 & \textbf{0.680} \\
TREx & P140 & 0.569 & 0.574 & \textbf{0.608} & 0.602 \\
TREx & P1412 & 0.764 & 0.773 & \textbf{0.785} & 0.781 \\
TREx & P159 & 0.535 & 0.551 & 0.573 & \textbf{0.576} \\
TREx & P17 & 0.870 & 0.863 & 0.884 & \textbf{0.887} \\
TREx & P176 & 0.647 & 0.673 & 0.699 & \textbf{0.720} \\
TREx & P178 & 0.569 & 0.592 & 0.631 & \textbf{0.639} \\
TREx & P19 & 0.477 & 0.478 & 0.509 & \textbf{0.519} \\
TREx & P190 & 0.279 & 0.276 & 0.296 & \textbf{0.297} \\
TREx & P20 & 0.511 & 0.533 & 0.559 & \textbf{0.565} \\
TREx & P264 & 0.247 & 0.280 & 0.291 & \textbf{0.313} \\
TREx & P27 & 0.745 & 0.756 & 0.767 & \textbf{0.773} \\
TREx & P276 & 0.625 & 0.623 & 0.652 & \textbf{0.663} \\
TREx & P279 & 0.512 & 0.544 & 0.562 & \textbf{0.580} \\
TREx & P30 & 0.802 & 0.813 & 0.835 & \textbf{0.842} \\
TREx & P31 & 0.616 & 0.627 & 0.635 & \textbf{0.635} \\
TREx & P36 & 0.569 & 0.578 & \textbf{0.618} & 0.615 \\
TREx & P361 & 0.530 & 0.538 & 0.567 & \textbf{0.574} \\
TREx & P364 & 0.703 & 0.715 & 0.729 & \textbf{0.742} \\
TREx & P37 & 0.701 & 0.688 & \textbf{0.728} & 0.715 \\
TREx & P39 & 0.572 & 0.607 & 0.613 & \textbf{0.630} \\
TREx & P407 & 0.638 & 0.630 & 0.647 & \textbf{0.666} \\
TREx & P413 & 0.422 & 0.453 & 0.483 & \textbf{0.507} \\
TREx & P449 & 0.416 & 0.444 & 0.454 & \textbf{0.495} \\
TREx & P463 & 0.646 & 0.674 & 0.697 & \textbf{0.713} \\
TREx & P47 & 0.492 & 0.508 & 0.564 & \textbf{0.565} \\
TREx & P495 & 0.685 & 0.662 & \textbf{0.699} & 0.681 \\
TREx & P527 & 0.423 & 0.452 & 0.521 & \textbf{0.527} \\
TREx & P530 & 0.379 & 0.373 & 0.400 & \textbf{0.416} \\
TREx & P740 & 0.407 & 0.414 & 0.438 & \textbf{0.438} \\
TREx & P937 & 0.528 & 0.541 & 0.569 & \textbf{0.569} \\
\bottomrule
\end{tabular}
}
\caption{Relation by relation performance comparison on LAMA~\cite{petroni-etal-2019-language}}
\end{table*}

\newpage
\section{Hyper-parameter for fine-tuning on GLUE, SQuAD}
\label{app:finetuning}

\begin{table}[!ht]
\centering
\resizebox{0.4\textwidth}{!}{
\begin{tabular}{ccc}
\toprule
Hyper-parameter & GLUE & SQuAD \\
\midrule
Batch Size & 32 & 12 \\
Learning Rate & 2e-5 & 3e-5 \\
Epochs & 3 & 2 \\
Weight Decay & 0.01 & 0.01 \\
\bottomrule
\end{tabular}
}
\caption{Fine-tuning hyper-parameters for GLUE and SQuAD}
\label{tab:fine_tuning_parameters}
\end{table}

\section{Hyper-parameter for AutoPrompt}
\label{app:autoprompt}

\begin{table}[!ht]
\centering
\begin{tabular}{ccc}
\toprule
Hyper-parameter & SST2 & NLI \\
\midrule
\# Trigger Token & 3 & 4 \\
\# Candidate & 100 & 10 \\
Batch Size & 24 & 32 \\
\# Iterations & 180 & 100\\
\bottomrule
\end{tabular}
\caption{Prompt-tuning hyper-parameters for AutoPrompt~\cite{shin-etal-2020-autoprompt}}
\label{tab:prompt_tuning_parameters}
\end{table}

\begin{table*}[!ht]
\resizebox{\textwidth}{!}{
\begin{tabular}{x{1cm}x{5cm}x{8cm}x{4cm}}
\toprule
Task & Template & Prompt Example & Labels \\
\midrule
SST2 & \{sentence\} \textcolor{red}{{[}T{]} . . . {[}T{]}} \textcolor{blue}{{[}P{]}} & director rob marshall went out gunning to make a great one \textcolor{red}{movie director cinema} \textcolor{blue}{{[}\texttt{MASK}{]}} &  \textbf{pos:} partnership, good  \textbf{neg:} worse, bad\\
\midrule
NLI & \{prem\}\textcolor{blue}{{[}P{]}} \textcolor{red}{{[}T{]} . . . {[}T{]}} \{hyp\} & There is no man in a black jacket doing tricks on a motorbike \textcolor{blue}{{[}\texttt{MASK}{]}} \textcolor{red}{strange workplace} A person in a black jacket is doing tricks on a motorbike & \textbf{con:} Nobody, nobody, nor \textbf{ent:} found, ways, Agency \textbf{neu:} \#\#ponents, \#\#lary, \#\#uated\\
\bottomrule
\end{tabular}
}
\caption{Prompt template for Sentiment Analysis and Natural Language Inference tasks in AutoPrompt~\cite{shin-etal-2020-autoprompt}}
\end{table*}